\documentclass[conference]{IEEEtran}
\IEEEoverridecommandlockouts
\usepackage{cite}
\usepackage{amsmath,amssymb,amsfonts}
\usepackage{algorithmic}
\usepackage{graphicx}
\usepackage{textcomp}
\usepackage{xcolor}
\usepackage{array}
\usepackage{verbatim}
\usepackage{siunitx}
\usepackage{rotating}
\usepackage[margin=1in]{geometry}
\usepackage{subcaption}
\usepackage{enumitem}
\usepackage{booktabs}
\usepackage{colortbl}
\usepackage{tabularx}
\usepackage{makecell}
\usepackage{pifont}
\usepackage{multirow}
\usepackage{microtype}
\usepackage[normalem]{ulem}
\usepackage{array}      % 提供 'm' 列类型，用于单元格内容的垂直居中
\usepackage[font=small,labelfont=bf]{caption}
\usepackage{arydshln}   % 提供在表格中绘制虚线的功能
\usepackage[dvipsnames,table]{xcolor}
\usepackage{tikz}
\usepackage{pifont}
\usetikzlibrary{positioning,calc}
\usetikzlibrary{arrows.meta}
\definecolor{dottedlineblue}{rgb}{0.6, 0.7, 0.9}
\PassOptionsToPackage{hyphens}{url}
\usepackage{array,colortbl}
\usepackage[
    colorlinks=true,
    linkcolor=blue,
    citecolor=green,
    urlcolor=red
]{hyperref}
\usepackage{cleveref}
\usepackage{nicematrix}
\crefname{figure}{Fig.}{Figs.}
\usepackage{dblfloatfix} % 导言区
\def\BibTeX{{\rm B\kern-.05em{\sc i\kern-.025em b}\kern-.08em
    T\kern-.1667em\lower.7ex\hbox{E}\kern-.125emX}}
\begin{document}

\title{Hierarchical Fusion of Local and Global Visual Features with Mixture-of-Experts for Remote Sensing Image Scene Classification}

\author{Yuanhao Tang, Xuechao Zou, Zhengpei Hu, Jianqiang Huang,Junliang Xing, Chengkun Zhang}

\maketitle

\begin{abstract}
Remote sensing image scene classification remains a challenging task, primarily due to the complex spatial structures and multi-scale characteristics of ground objects. Although CNN-based methods excel at extracting local inductive biases, and Mamba-based approaches demonstrate impressive capabilities in efficiently capturing global sequential context, relying on a single paradigm restricts the model's ability to simultaneously characterize fine-grained textures and complex spatial structures. To tackle this, we propose a parallel heterogeneous encoder, a hierarchical fusion module designed to achieve effective local-global co-representation. It consists of two parallel pathways: a local visual encoder for extracting multi-scale local visual features, and a global visual encoder for capturing efficient global visual features. The core innovation lies in its hierarchical fusion module, which progressively aggregates multi-scale features from both pathways, enabling dynamic cross-level feature interaction and contextual reconstruction to produce highly discriminative representations. These fused features are then adaptively routed through a mixture-of-experts classifier head, which dynamically dispatches them to the most suitable experts for fine-grained scene recognition. Experiments on AID, NWPU-RESISC45, and UC Merced show that our model achieves 93.72\%, 95.54\%, and 96.92\% accuracy, surpassing SOTA methods with an optimal balance of performance and efficiency. Code is available at \url{https://anonymous.4open.science/r/classification-41DF}.
\end{abstract}


\section{INTRODUCTION}
\label{sec:intro}

\begin{figure}[t]
	\centering
	\includegraphics[width=\linewidth]{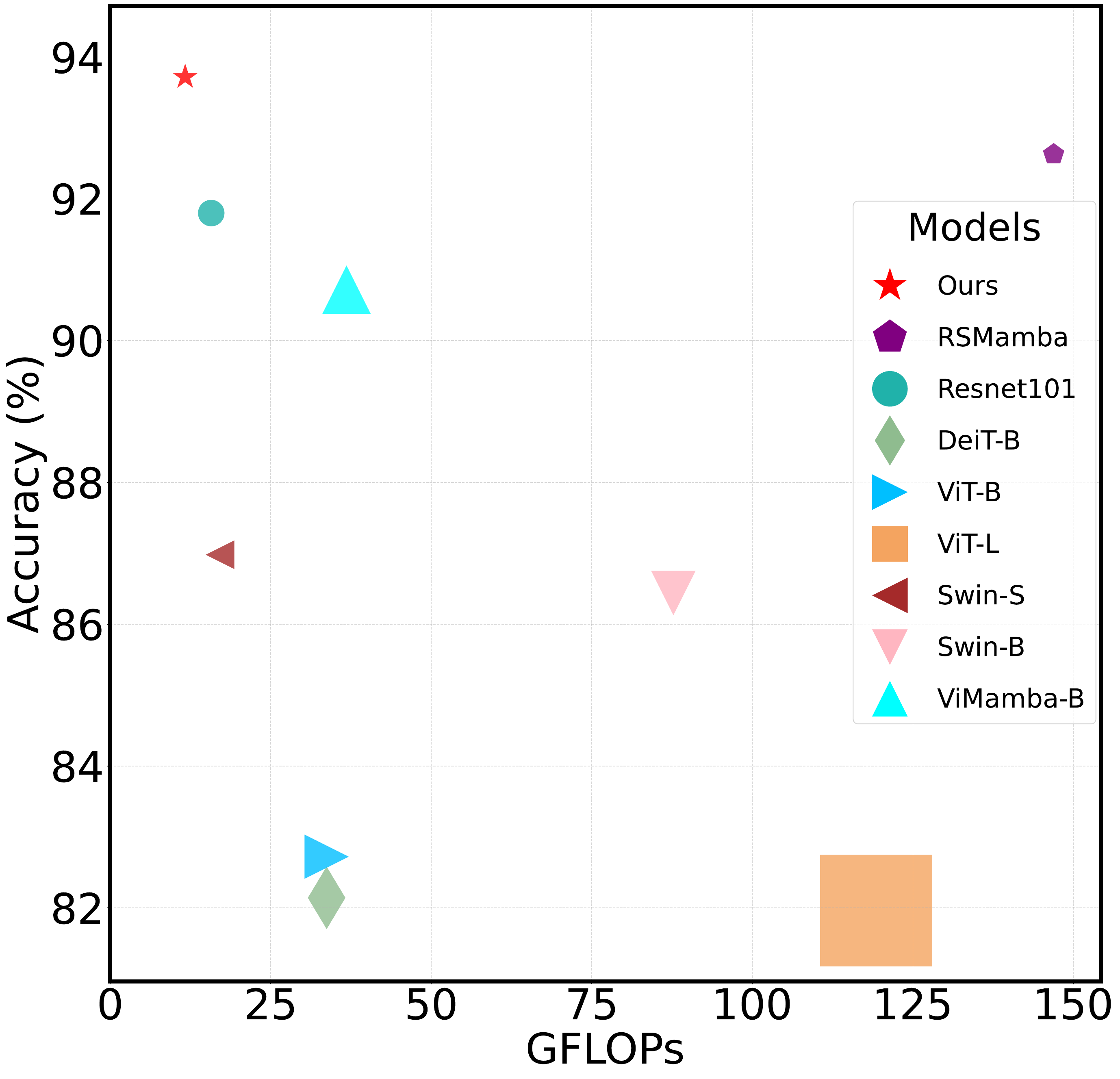}
	\caption{Accuracy and efficiency comparison. Bubble size denotes parameters. Our method (red star) achieves the optimal balance.}
	\label{fig:complexity_vs_accuracy}
\end{figure}

\IEEEPARstart{W}{ith} the rapid advancement of Earth observation technologies, high-resolution remote sensing imagery has become increasingly accessible~\cite{ref1}. 
Covering continuous spectral bands~\cite{ref2}, such imagery provides rich spatial and semantic information for applications such as urban mapping and environmental monitoring~\cite{ref6,aid}. 
Remote sensing image scene classification (RSIC) aims to assign semantic labels to aerial scenes but remains challenging due to complex spatial structures, large-scale variations, and high intra-class diversity.

Recent RSIC methods are dominated by deep learning approaches~\cite{adegun2023review}, mainly including convolutional neural networks (CNNs)~\cite{castelluccio2015land} and Transformer-based models~\cite{zhang2021trs,bazi2021vision}. 
CNNs excel at modeling local textures and structures but struggle with long-range spatial interactions, while Transformers capture global context at the cost of quadratic complexity and may overlook fine-grained local cues.

\begin{figure*}[t!]
    \centering
    \begin{subfigure}[b]{0.32\textwidth}
        \centering
        \includegraphics[width=\linewidth]{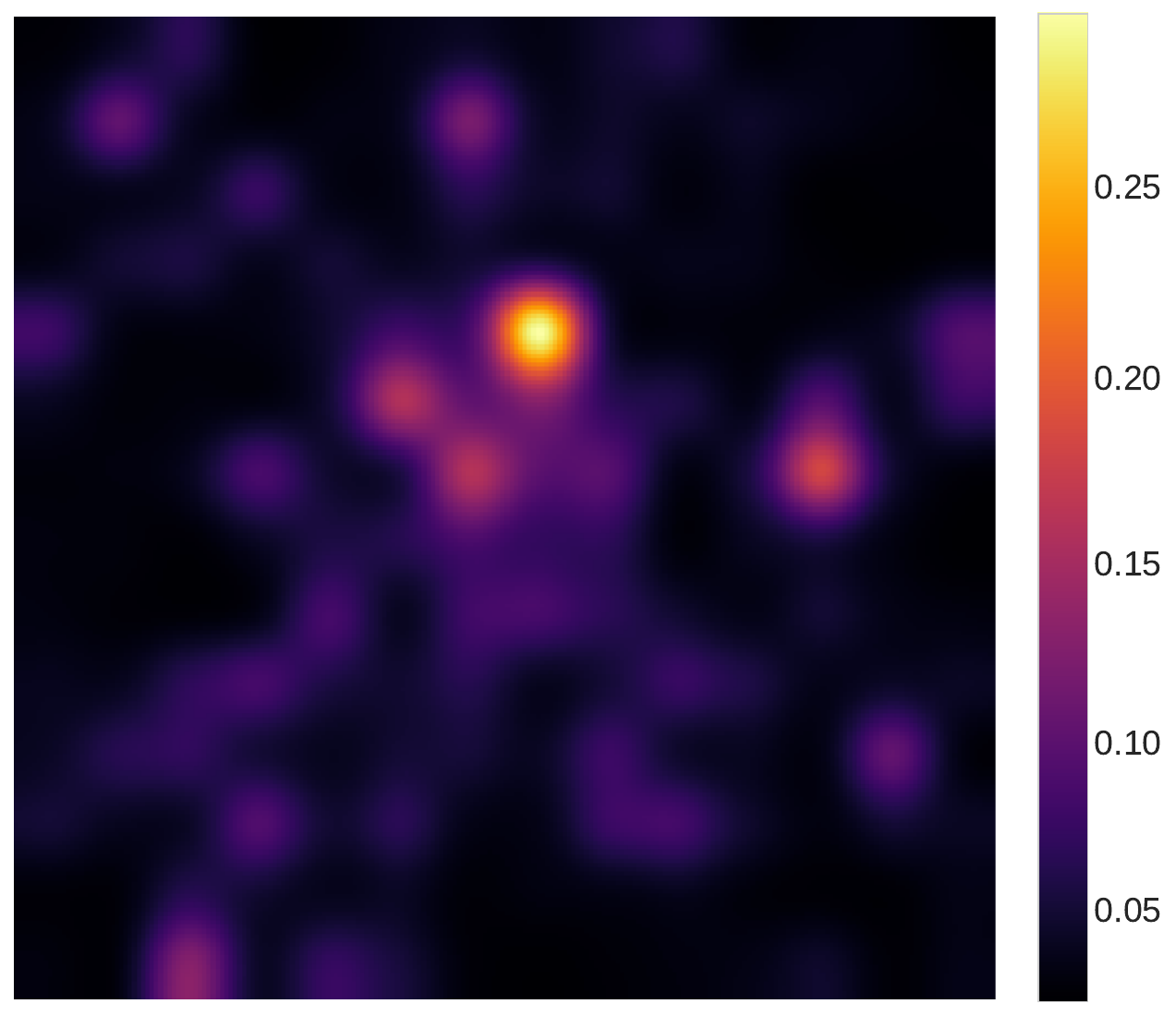}
        \caption{Transformer}
    \end{subfigure}
    \hfill
    \begin{subfigure}[b]{0.32\textwidth}
        \centering
        \includegraphics[width=\linewidth]{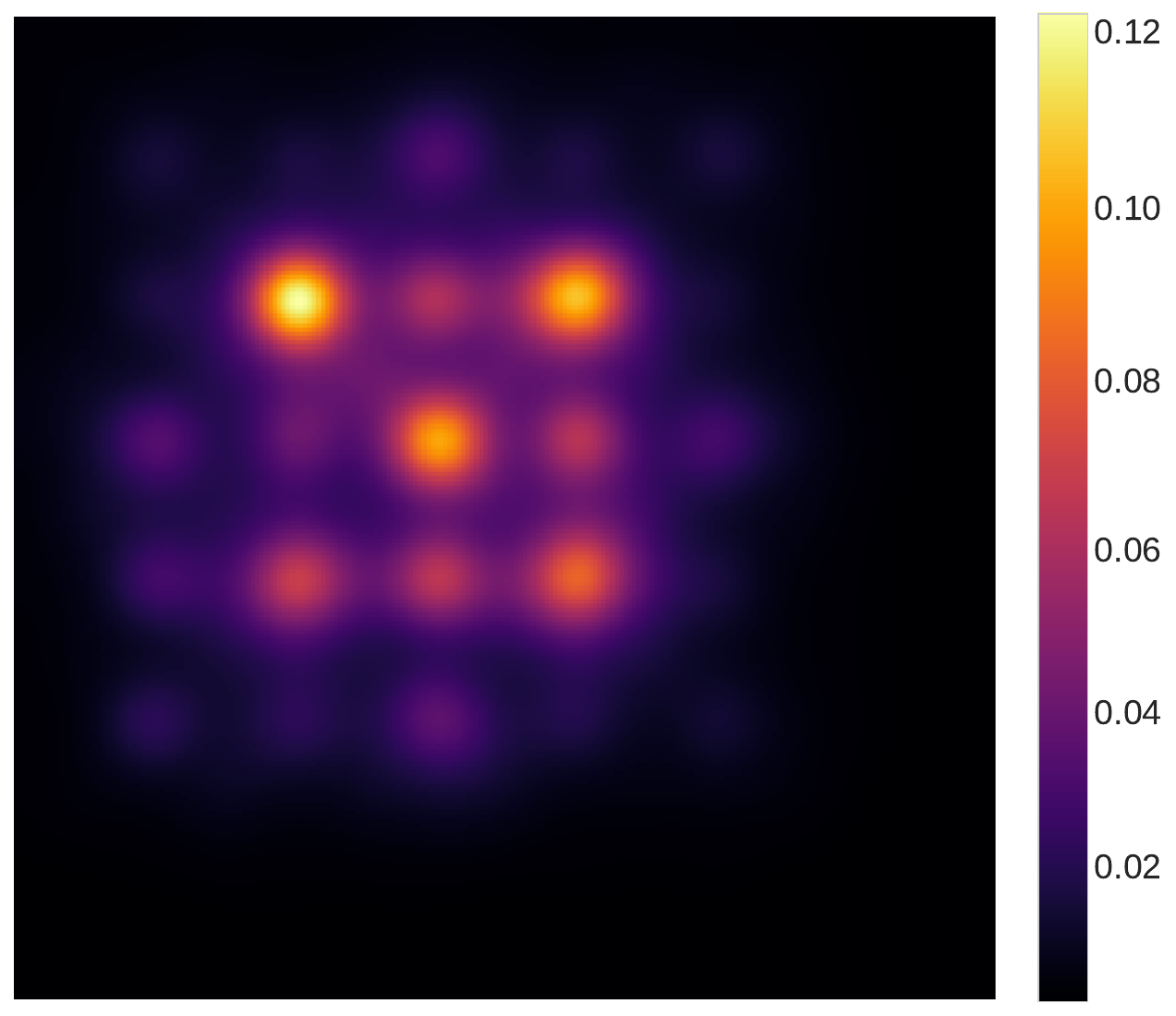}
        \caption{CNN}
    \end{subfigure}
    \hfill
    \begin{subfigure}[b]{0.32\textwidth}
        \centering
        \includegraphics[width=\linewidth]{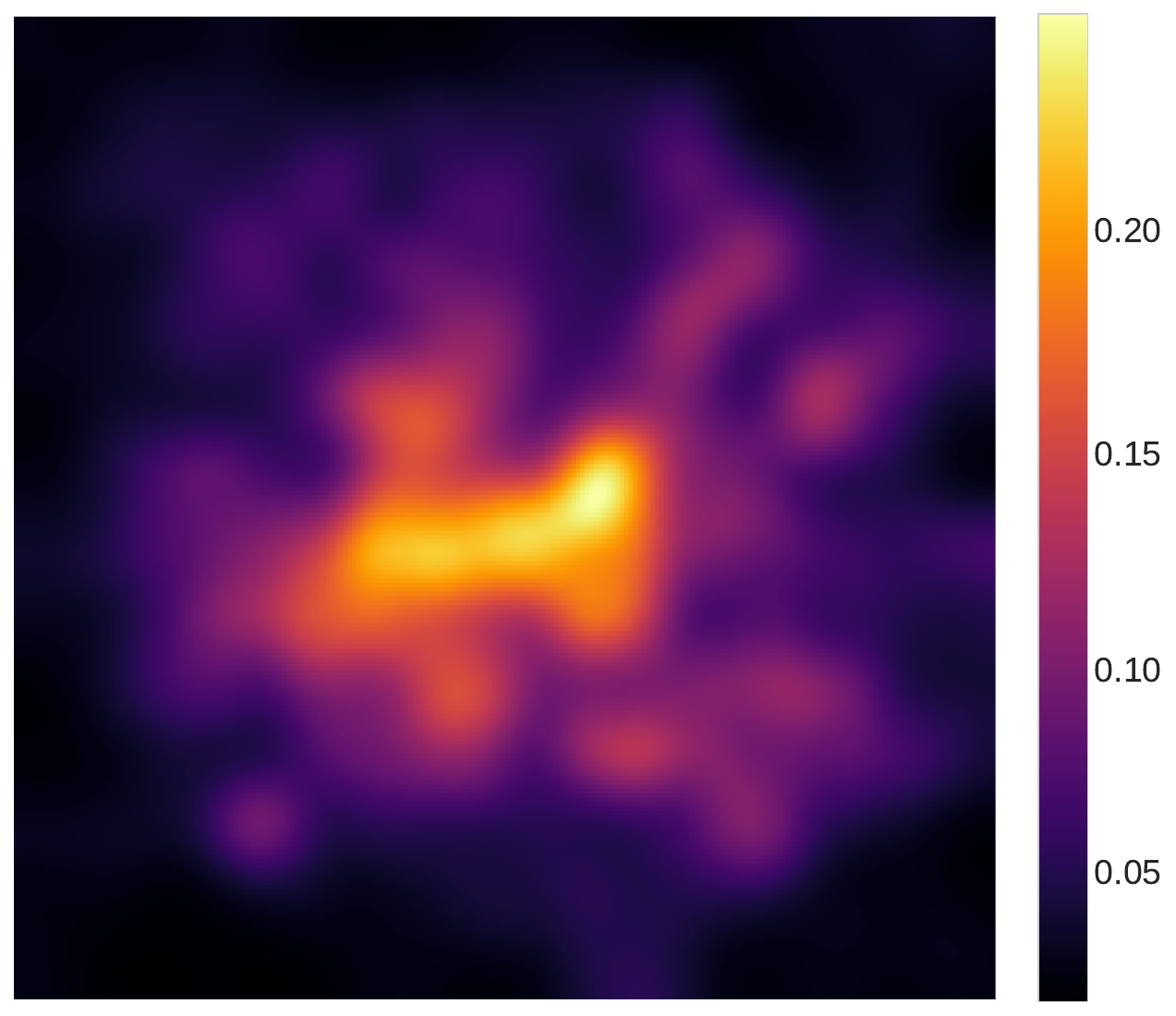}
        \caption{Ours}
    \end{subfigure}
    \caption{Effective receptive field (ERF) comparison~\cite{ding2022scaling}.}
    \label{fig:erf_comparison}
\end{figure*}

Recently, Mamba~\cite{gu2023mamba}, built upon state space models (SSMs)~\cite{gu2023mamba}, has emerged as an efficient alternative for long-sequence modeling with linear complexity and has been extended to vision tasks~\cite{zhu2024vision}. 
However, existing RSIC approaches often adopt Mamba as a standalone backbone, implicitly assuming that stronger global modeling alone is sufficient, while neglecting the importance of strong local inductive priors.

In contrast, we argue that effective remote sensing scene understanding does not require a heavy global modeling backbone, but rather a lightweight and efficient mechanism to inject global contextual priors into locally discriminative representations. 
Motivated by this insight, we propose a parallel heterogeneous encoder that explicitly decouples local and global representation learning while maintaining high computational efficiency (see \Cref{fig:complexity_vs_accuracy}). 
Specifically, a CNN-based local encoder extracts multi-scale local textures and structural cues, whereas a shallow Mamba-based encoder is deliberately designed as a lightweight global context enhancer, capturing long-range dependencies without overwhelming local feature discrimination.

To bridge the semantic gap between heterogeneous representations, we design a hierarchical fusion module with densely connected dual-attention multi-scale fusion blocks, enabling progressive cross-level feature interaction. 
Furthermore, to explicitly handle the high intra-class variance inherent in remote sensing scenes, we introduce a mixture-of-experts (MoE) classifier head~\cite{liu2024deepseek}, which enables conditional specialization through dynamic routing with minimal overhead.

The main contributions of this paper are summarized as follows:
\begin{itemize}
    \item We propose a parallel heterogeneous encoder that synergizes CNN-based local inductive priors with a lightweight Mamba-based global context enhancer for efficient local--global co-representation.
    \item We design a hierarchical fusion module with dense cross-scale connections to align heterogeneous features and facilitate progressive semantic interaction.
    \item We introduce a task-aware mixture-of-experts classifier head to address high intra-class variance in remote sensing scene classification.
    \item Extensive experiments on three benchmarks demonstrate state-of-the-art performance with a favorable accuracy--efficiency trade-off.
\end{itemize}

\section{RELATED WORK}
\label{sec:related_work}

\subsection{Remote Sensing Image Scene Classification}
The design of RSIC models has evolved to solve the trade-off between receptive field and efficiency. (1) CNNs \cite{he2016deep} lead local feature extraction but are restricted by local receptive fields, limiting long-range dependency modeling. (2) vision transformers (ViTs) \cite{dosovitskiy2020image,liu2021swin} capture global context via self-attention but incur quadratic complexity, prompting hybrid designs \cite{wu2023cmtfnet}. (3) Recently, Mamba \cite{gu2023mamba,li2024spmamba} yields global modeling with linear complexity via selective scanning. This paradigm is now used for remote sensing \cite{chen2024rsmamba}, offering a robust solution to the limitations of prior arts.

\subsection{Mixture-of-Experts}
MoE \cite{jacobs1991adaptive,togo} increases model capacity via conditional computation, where a gating network dynamically selects specialized experts for each input. MoE has demonstrated strong scalability and performance in natural language processing and computer vision tasks \cite{riquelme2021scaling}. In the remote sensing domain, however, its applications remain relatively limited, primarily focusing on image captioning or multi-task learning scenarios \cite{lin2025rs}.

\section{METHOD}
\label{sec:method}
\begin{figure*}[!t]
	\centering
	\includegraphics[width=1\textwidth]{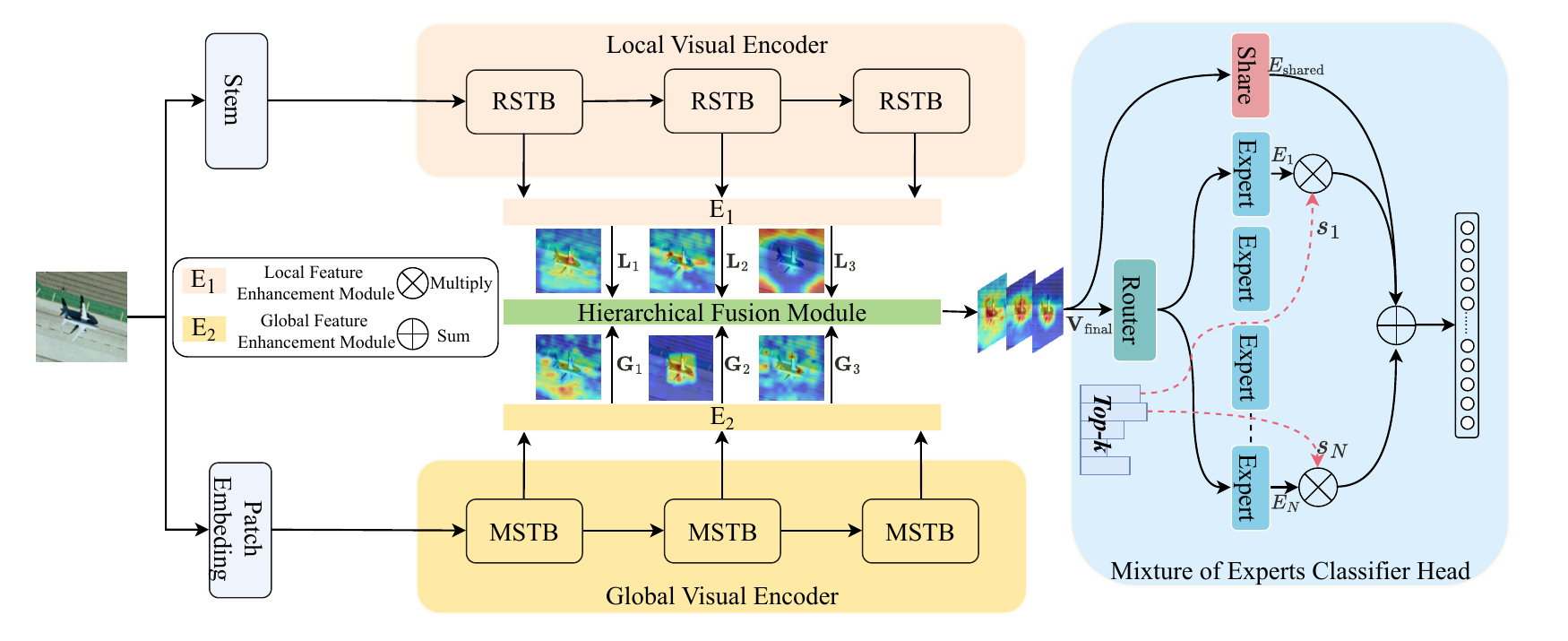}
	\caption{Architecture of the proposed model. It comprises a parallel local-global visual encoder, a hierarchical fusion module, and a mixture-of-experts classifier head.}
	\label{Fig1}
\end{figure*}
We propose a framework to integrate local details with global semantics. As shown in \Cref{Fig1}, it comprises a parallel heterogeneous encoder, a hierarchical fusion module, and a mixture-of-experts (MoE) head. First, the local and global visual encoders extract visual features and global sequences, respectively. Next, the central hierarchical fusion module aggregates the extracted features, followed by an MoE head for adaptive classification. Overall, this approach effectively leverages local-global synergy and expert-based decision-making.

\subsection{Parallel Heterogeneous Encoder}
As illustrated in \Cref{Fig1}, the proposed architecture employs a parallel heterogeneous encoder to extract complementary visual information through two specialized branches: the local visual encoder and the global visual encoder.

\subsubsection{Local Visual Encoder}
Built on ResNet, this branch focuses on learning fine-grained local textures. After a Stem module reduces spatial redundancy, features pass through stacked ResNet stage blocks (RSTB). While a standard ResNet backbone traditionally consists of four stages, we selectively extract features from the last three stages to strike an optimal balance between semantic depth and computational cost. These hierarchical outputs are further refined by the local feature enhancement module ($\mathbf{E}_{1}$) to align their distributions, yielding the enhanced local features $\mathbf{L} = \{ \mathbf{L}_i \in \mathbb{R}^{C_i \times H_i \times W_i} \}_{i=1}^3$.

\subsubsection{Global Visual Encoder}
Adopting the Vision Mamba backbone, this branch models global context. The hierarchical features are modeled by cascaded mamba stage blocks (MSTB), which leverage SSM to capture global semantic information with linear complexity. Following spatial reshaping to restore 2D structure, the features are further refined by the global feature enhancement module ($\mathbf{E}_{2}$). This process yields the enhanced global features $\mathbf{G} = \{ \mathbf{G}_i \in \mathbb{R}^{C_i \times H_i \times W_i} \}_{i=1}^3$, serving as the global semantic guidance for the subsequent fusion module.

\subsection{Hierarchical Fusion Module}
\label{sec:dense_fusion}
\begin{figure}[t!] % 改为单栏图
    \centering
    % 将宽度适当调小（例如 0.45~0.5\textwidth 更适合单栏）
    \includegraphics[width=0.48\textwidth]{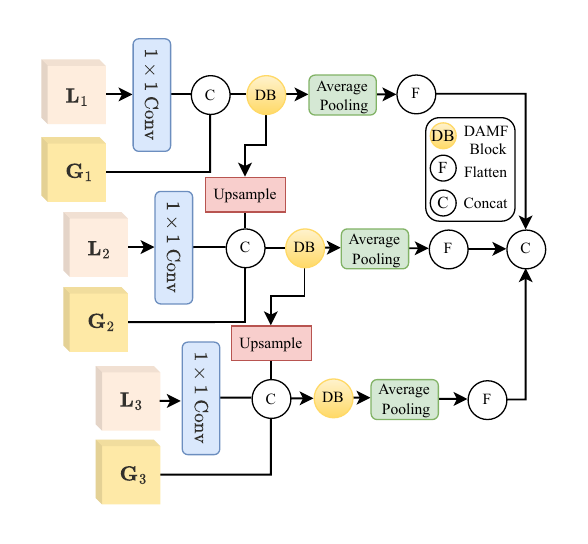}
    
    \caption{Details of the proposed hierarchical fusion module.}
    
    \label{fig:fusion_module}
\end{figure}

The hierarchical fusion module (\Cref{fig:fusion_module}) bridges local and global branches via a top-down aggregation strategy in three steps:

First, local features $\mathbf{L}_{i}$ are aligned via a $1\times1$ convolution: $\mathbf{\hat{L}}_{i} = \text{Conv}_{1\times1}(\mathbf{L}_{i})$. 

Second, the aligned $\mathbf{\hat{L}}_{i}$ and global features $\mathbf{{G}}_{i}$ are fused with upsampled priors from the previous stage to generate the refined feature map $\mathbf{Z}_{i} \in \mathbb{R}^{C_i \times H_i \times W_i}$:
\begin{equation}
    \mathbf{{Z}}_{i} = 
    \begin{cases} 
    \text{DB}([\mathbf{\hat{L}}_{i}, \mathbf{{G}}_{i}]), & i=1 \\
    \text{DB}([\mathbf{\hat{L}}_{i}, \mathbf{{G}}_{i}, \text{Up}(\mathbf{{Z}}_{i-1})]), & i \in \{2, 3\}
    \end{cases}
\end{equation}
where $[\cdot]$, $\text{Up}(\cdot)$, and \text{DB} denote concatenation, upsampling, and the DAMF-Block, respectively.

Finally, the refined features $\mathbf{{Z}}_{i}$ are pooled and flattened into vectors $\mathbf{v_i} \in \mathbb{R}^{C_i}$, which are concatenated into the final representation:
\begin{equation}
    \mathbf{V}_{\text{final}} = [\mathbf{v_1}, \mathbf{v_2}, \mathbf{v_3}] \in \mathbb{R}^{C_{\text{total}}},
\end{equation}
where $C_{\text{total}} = \sum C_i$. $\mathbf{V}_{\text{final}}$ effectively suppresses background noise and serves as the input for the classifier.

\begin{table*}[t!]
    \centering
    \scriptsize
    \renewcommand{\arraystretch}{1.1} % 稍微增加行间距，看起来不拥挤
    \caption{Sota performance conparison on three datasets (Trained from scratch).}
    \label{tab:performance_comparison}

    % 使用 tabular 并通过 @{\hskip 宽度} 自定义大组之间的间距
    \begin{tabular}{
            l 
            @{\hskip 40pt} S[table-format=2.2] S[table-format=2.2] S[table-format=2.2] % UC Merced
            @{\hskip 40pt} S[table-format=2.2] S[table-format=2.2] S[table-format=2.2] % AID
            @{\hskip 40pt} S[table-format=2.2] S[table-format=2.2] S[table-format=2.2] % NWPU
        }
        \toprule
        \textbf{Model} & \multicolumn{3}{c}{\textbf{UC Merced (\%)}} & \multicolumn{3}{c}{\textbf{AID (\%)}} & \multicolumn{3}{c}{\textbf{NWPU (\%)}} \\
        \cmidrule(lr){2-4} \cmidrule(lr){5-7} \cmidrule(lr){8-10} 
        & {P} & {R} & {F1} & {P} & {R} & {F1} & {P} & {R} & {F1} \\
        \midrule
        \multicolumn{10}{l}{\textit{CNN-based Models}} \\ 
        ResNet18\cite{he2016deep}    & 90.40 & 90.32 & 90.22 & 92.28 & 92.24 & 92.22 & 93.82 & 93.75 & 93.75 \\
        ResNet50\cite{he2016deep}    & 91.25 & 90.95 & 90.85 & 92.34 & 92.26 & 92.22 & 94.44 & 94.40 & 94.40 \\
        ResNet101\cite{he2016deep}   & 93.93 & 93.81 & 93.74 & 91.87 & 91.80 & 91.77 & 94.80 & 94.75 & 94.75 \\
        \midrule
        \multicolumn{10}{l}{\textit{Transformer-based Models}} \\
        DeiT-T\cite{touvron2021training}   & 85.07 & 84.44 & 84.42 & 80.74 & 80.72 & 80.63 & 83.57 & 83.57 & 83.45 \\
        DeiT-S\cite{touvron2021training}   & 91.90 & 91.75 & 91.61 & 80.98 & 81.04 & 80.92 & 83.12 & 82.99 & 82.98 \\
        DeiT-B\cite{touvron2021training}   & 92.53 & 92.38 & 92.34 & 82.20 & 82.14 & 82.05 & 80.11 & 80.08 & 79.98 \\
        \cmidrule(lr){1-10}
        ViT-B\cite{dosovitskiy2020image}    & 90.49 & 90.32 & 90.18 & 82.79 & 82.72 & 82.54 & 80.25 & 80.26 & 80.16 \\
        ViT-L\cite{dosovitskiy2020image}    & 92.80 & 92.54 & 92.42 & 82.08 & 81.96 & 81.84 & 79.50 & 79.56 & 79.46 \\
        \cmidrule(lr){1-10}
        Swin-T\cite{liu2021swin}   & 89.28 & 88.89 & 88.88 & 87.41 & 87.40 & 87.35 & 89.79 & 89.75 & 89.72 \\
        Swin-S\cite{liu2021swin}   & 90.01 & 89.84 & 89.72 & 87.03 & 86.98 & 87.03 & 89.42 & 89.28 & 89.28 \\
        Swin-B\cite{liu2021swin}   & 91.93 & 91.75 & 91.66 & 86.51 & 86.44 & 86.37 & 89.55 & 89.40 & 89.41 \\
        CMTFNet\cite{wu2023cmtfnet}   & 94.80 & 95.04 & 94.64 & 93.23 & 93.01 & 93.01 & 94.45 & 94.35 & 94.31 \\
        \midrule
        \multicolumn{10}{l}{\textit{Mamba-based Models}} \\
        VMamba-T\cite{liu2024vmamba}       & 93.14 & 92.85 & 92.81 & 91.59 & 90.94 & 91.10 & 93.97 & 93.96 & 93.94 \\
        Vision Mamba-T\cite{zhu2024vision}   & 83.83 & 83.81 & 83.06 & 79.16 & 78.94 & 78.68 & 89.24 & 89.02 & 88.97 \\
        Vision Mamba-S\cite{zhu2024vision}   & 89.62 & 89.68 & 89.32 & 87.77 & 87.66 & 87.54 & 95.23 & 95.22 & 95.21 \\
        Vision Mamba-B\cite{zhu2024vision}   & 88.94 & 89.05 & 88.82 & 90.98 & 90.80 & 90.72 & 95.10 & 95.07 & 95.06 \\
        \cmidrule(lr){1-10}
        RSMamba-B\cite{chen2024rsmamba}      & 94.14 & 93.97 & 93.88 & 92.02 & 91.53 & 91.66 & 94.87 & 94.87 & 94.84 \\
        RSMamba-L\cite{chen2024rsmamba}      & 95.03 & 94.76 & 94.74 & 92.31 & 91.75 & 91.90 & 95.03 & 95.05 & 95.02 \\
        RSMamba-H\cite{chen2024rsmamba}      & 95.47 & 95.23 & 95.25 & 92.97 & 92.51 & 92.63 & 95.22 & 95.19 & 95.18 \\
        \cmidrule(lr){1-10}
        HC-Mamba-T\cite{yang2025hc}         & 94.12 & 94.59 & 94.76 & 91.97 & 91.47 & 91.42 & 94.88 & 94.96 & 94.87 \\
        HC-Mamba-S\cite{yang2025hc}         & 95.10 & 95.00 & 95.08 & 92.33 & 91.88 & 91.95 & 95.10 & 95.12 & 95.08 \\
        HC-Mamba-B\cite{yang2025hc}         & 95.55 & 95.31 & 95.34 & 93.02 & 92.68 & 92.86 & 95.32 & 95.26 & 95.25 \\
        \midrule[1pt]
        \multicolumn{1}{l}{\bfseries Ours} & 
        {\bfseries 96.92} & {\bfseries 96.83} & {\bfseries 96.81} & 
        {\bfseries 93.76} & {\bfseries 93.72} & {\bfseries 93.71} & 
        {\bfseries 95.54} & {\bfseries 95.52} & {\bfseries 95.52} \\
        \bottomrule
    \end{tabular}
\end{table*}

\subsection{Mixture-of-Experts Classifier Head}
To efficiently address high intra-class variance, we employ a MoE head (right panel of \Cref{Fig1})\cite{liu2024deepseek}. It comprises a fixed \textit{shared expert} for common patterns and $N$ \textit{routed experts} with dynamic Top-$k$ activation. Given the input $\mathbf{V}_{\text{final}}$, a router generates gating scores $\mathbf{S} \in \mathbb{R}^{N}$. The output $\mathbf{V}_{\text{out}} \in \mathbb{R}^{C_{total}}$ sums the shared and weighted selected features:
\begin{equation}
\resizebox{0.9\linewidth}{!}{$
\mathbf{V}_{\text{out}} = \underbrace{E_{\text{shared}}(\mathbf{V}_{\text{final}})}_{\text{Shared Knowledge}} + \underbrace{\sum_{j \in \text{Top-}k(\mathbf{S})} s_j E_j(\mathbf{V}_{\text{final}})}_{\text{Specialized Knowledge}},
$}
\end{equation}
where $E_{\text{shared}}$ and $E_j$ denote expert functions, and $s_j$ is the normalized weight from $\mathbf{S}$. To prevent \textit{expert collapse}, we add an auxiliary load-balancing loss $\mathcal{L}_{\text{aux}}$. The total objective is:
\begin{equation}
    \mathcal{L}_{\text{total}} = \mathcal{L}_{\text{CE}}(\mathbf{y}, \hat{\mathbf{y}}) + \lambda \mathcal{L}_{\text{aux}},
\end{equation}
where $\mathcal{L}_{\text{CE}}$ computes the cross-entropy between the ground truth $\mathbf{y}$ and the prediction $\hat{\mathbf{y}}$ (with $\mathbf{y}, \hat{\mathbf{y}} \in \mathbb{R}^{N_{cls}}$), and we set $\lambda = 0.01$ to regulate the trade-off. $\mathcal{L}_{\text{aux}}$ penalizes routing imbalance.

\section{EXPERIMENT}
\label{sec:experiments}

\subsection{Datasets}
We evaluate the proposed method on three datasets: UC Merced~\cite{ref6}, AID~\cite{aid}, and NWPU-RESISC45~\cite{ref7}. Varying significantly in scale, resolution, and data volume, these datasets provide a comprehensive assessment of the model's robustness and generalization capability across diverse scene complexities.

\subsection{Implementation Details}
All models are implemented in PyTorch and trained from scratch for 500 epochs on NVIDIA A800 GPUs. We utilize the AdamW \cite{loshchilov2018decoupled} optimizer with a batch size of 256, an initial learning rate of $5\times10^{-4}$, and weight decay of 0.05, modulated by a cosine annealing schedule with linear warm-up. Input images are resized to $224 \times 224$ with standard data augmentation (random flipping and color jittering). The objective function is cross-entropy loss with label smoothing ($\epsilon=0.1$). 
For evaluation, we report accuracy alongside weighted precision, recall, and F1-score to assess performance.

\begin{figure*}[t!]
    \centering
    % --- 子图 (a): UC ---
    \begin{subfigure}[t]{0.32\textwidth}
        \centering
        \includegraphics[width=\linewidth]{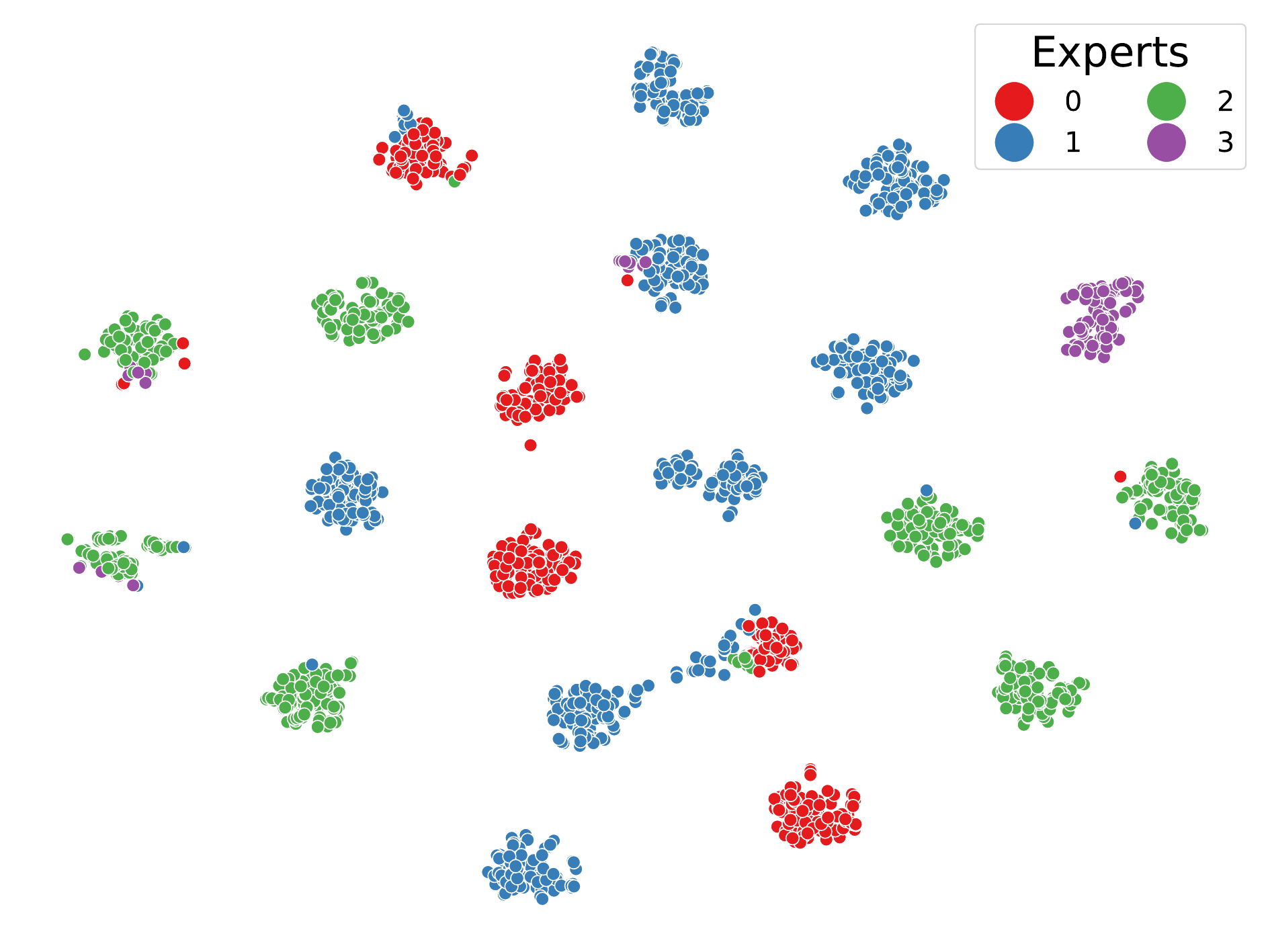}
        \caption{UC Merced}
        \label{fig:tsne_ucm}
    \end{subfigure}%
    \hfill
    % --- 子图 (b): AID ---
    \begin{subfigure}[t]{0.32\textwidth}
        \centering
        \includegraphics[width=\linewidth]{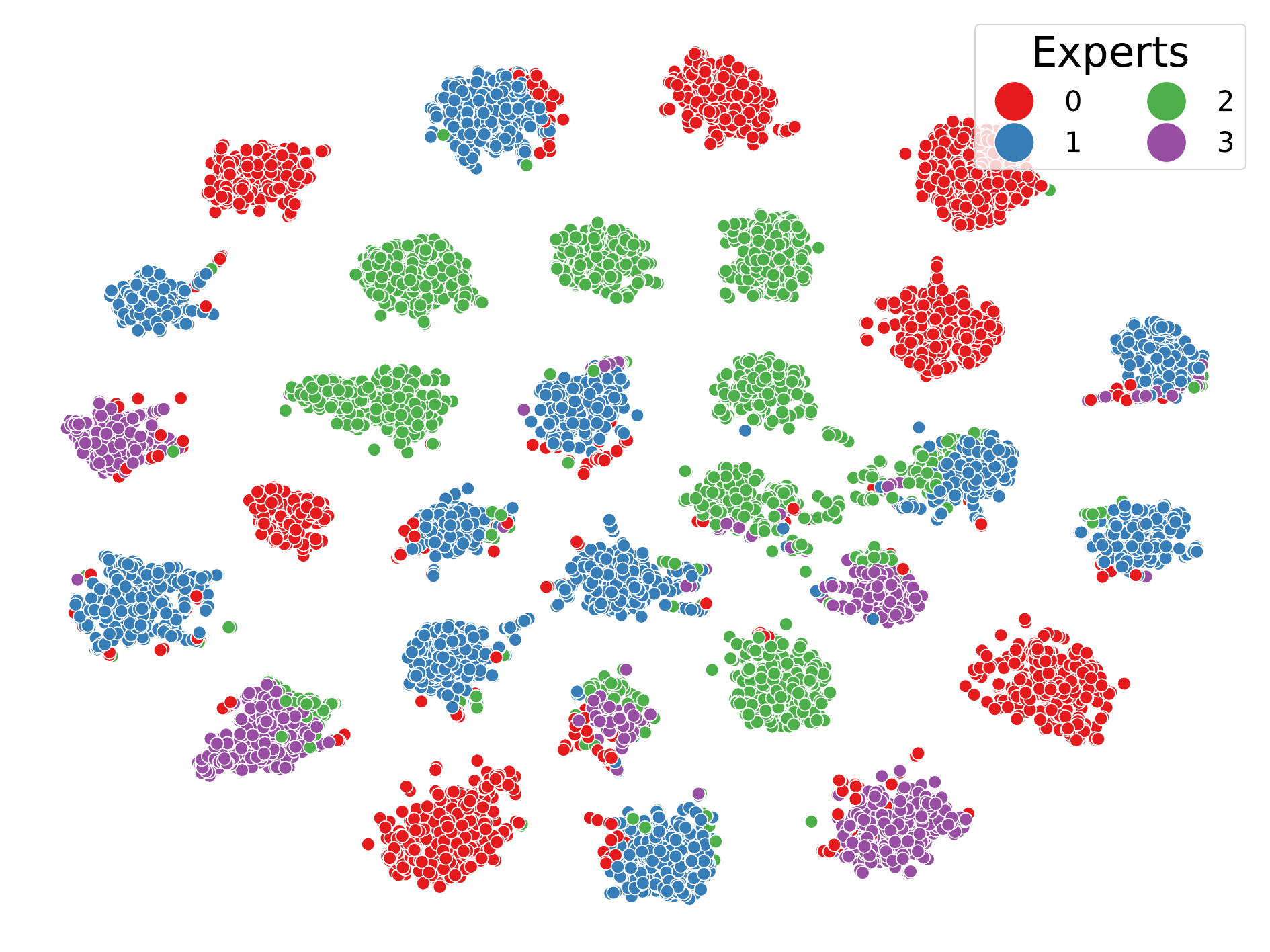}
        \caption{AID}
        \label{fig:tsne_aid}
    \end{subfigure}%
    \hfill
    % --- 子图 (c): NWPU ---
    \begin{subfigure}[t]{0.32\textwidth}
        \centering
        \includegraphics[width=\linewidth]{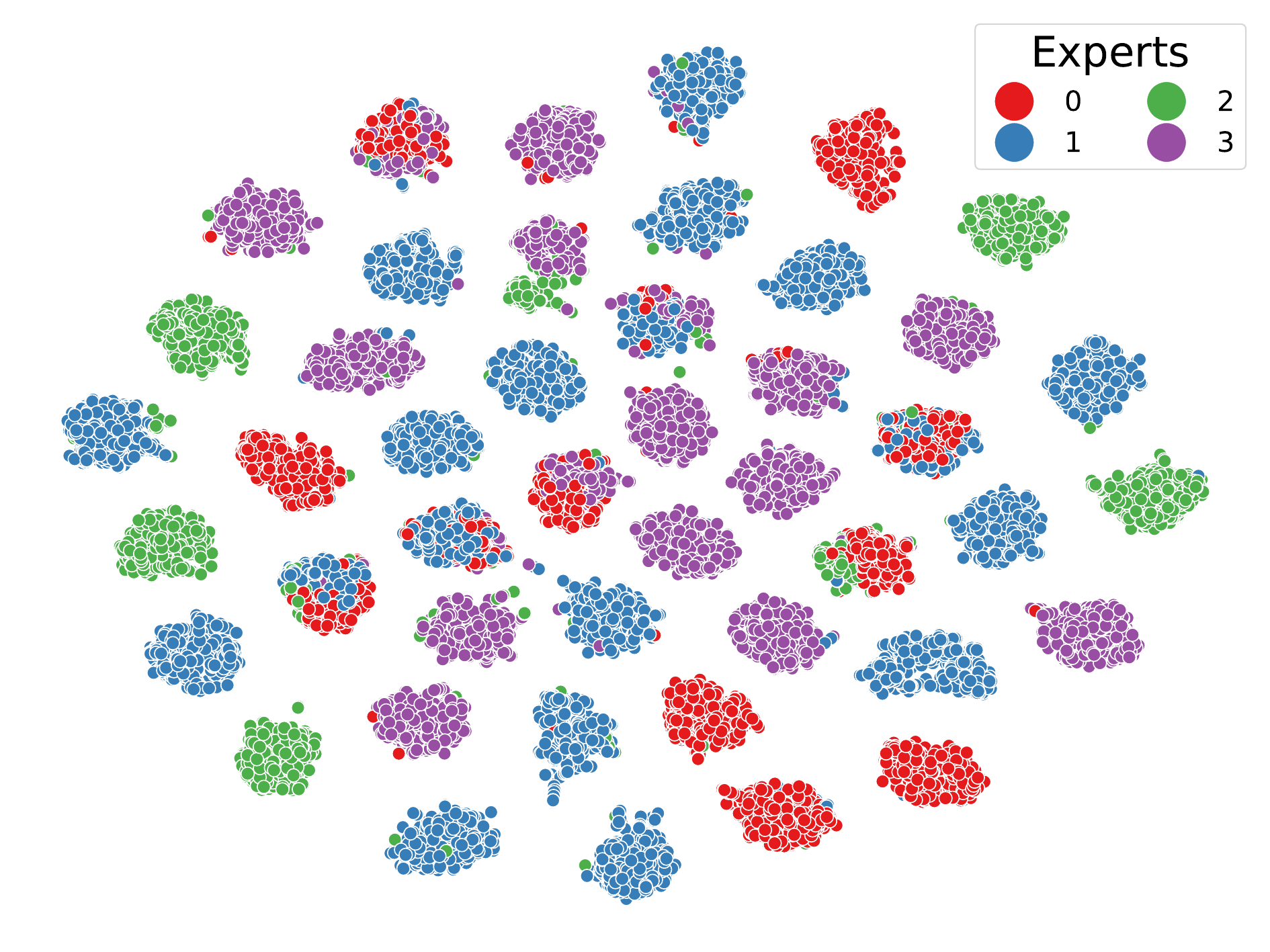}
        \caption{NWPU-RESISC45}
        \label{fig:tsne_nwpu}
    \end{subfigure}
    
    \caption{
t-SNE \cite{van2008visualizing} colored by dominant expert. Distinct clusters confirm structured routing.}
    \label{fig:tsne_visualization}
\end{figure*}
\begin{figure*}[t!]
    \centering
    % --- 子图 (a) ---
    \begin{subfigure}[t]{0.32\textwidth}
        \centering
        \includegraphics[width=\linewidth]{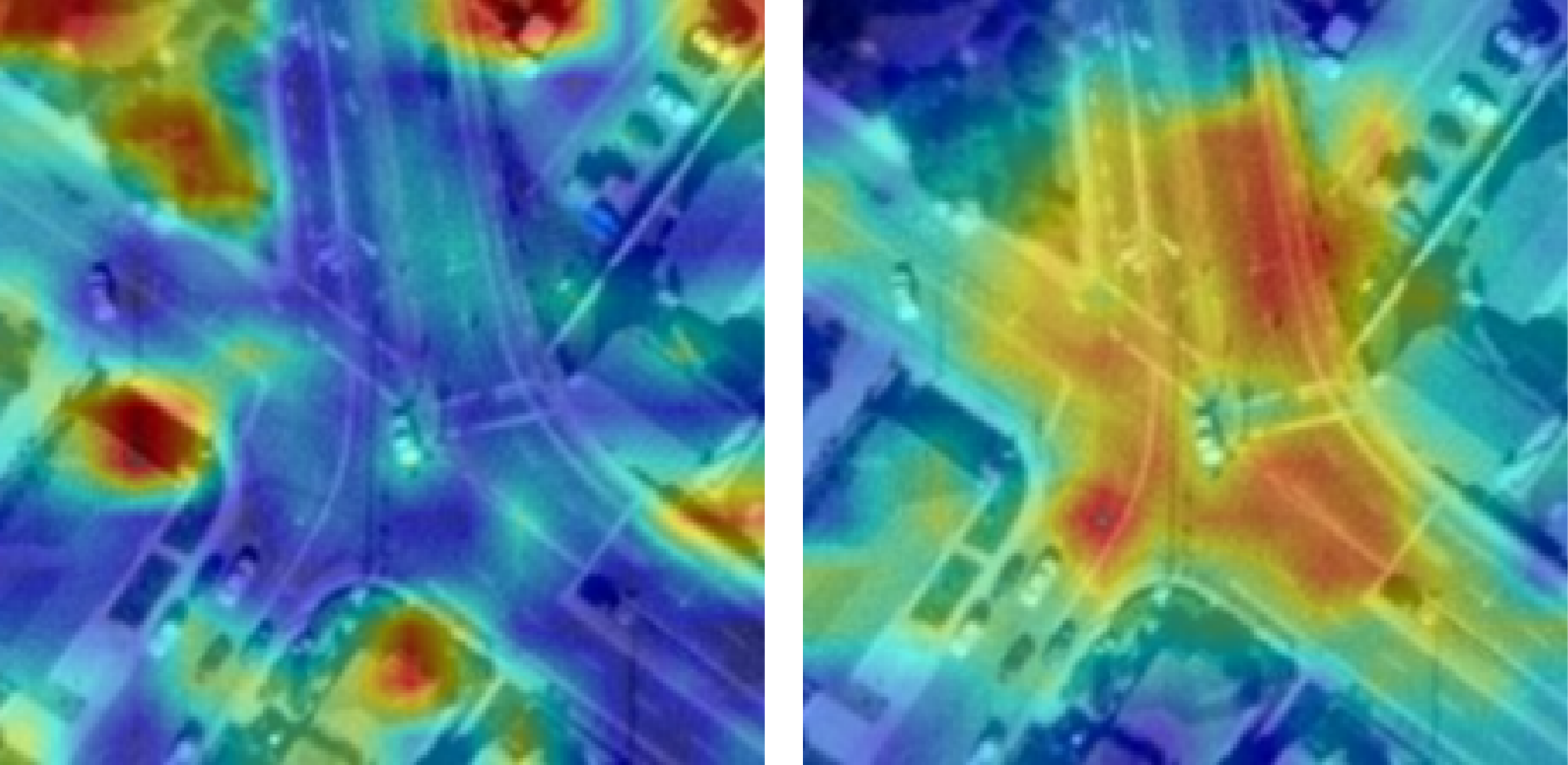}
        \caption{“Intersection” scene (UC Merced)}
        \label{fig:cam_harbor}
    \end{subfigure}%  <- 注意这里的 %
    \hfill
    % --- 子图 (b) ---
    \begin{subfigure}[t]{0.32\textwidth}
        \centering
        \includegraphics[width=\linewidth]{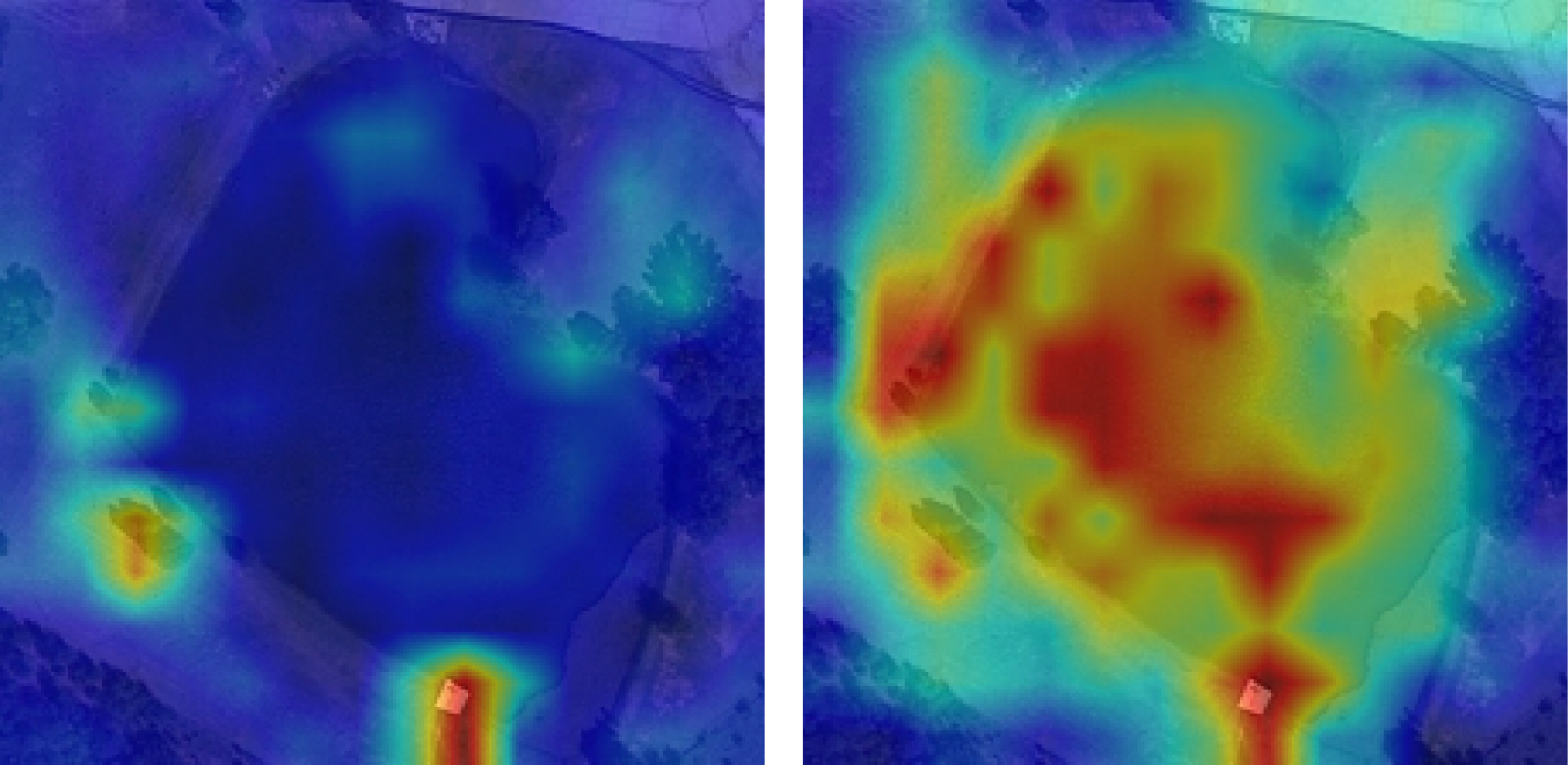}
        \caption{“Pond” scene (AID)}
        \label{fig:cam_church}
    \end{subfigure}%  <- 注意这里的 %
    \hfill
    % --- 子图 (c) ---
    \begin{subfigure}[t]{0.32\textwidth}
        \centering
        \includegraphics[width=\linewidth]{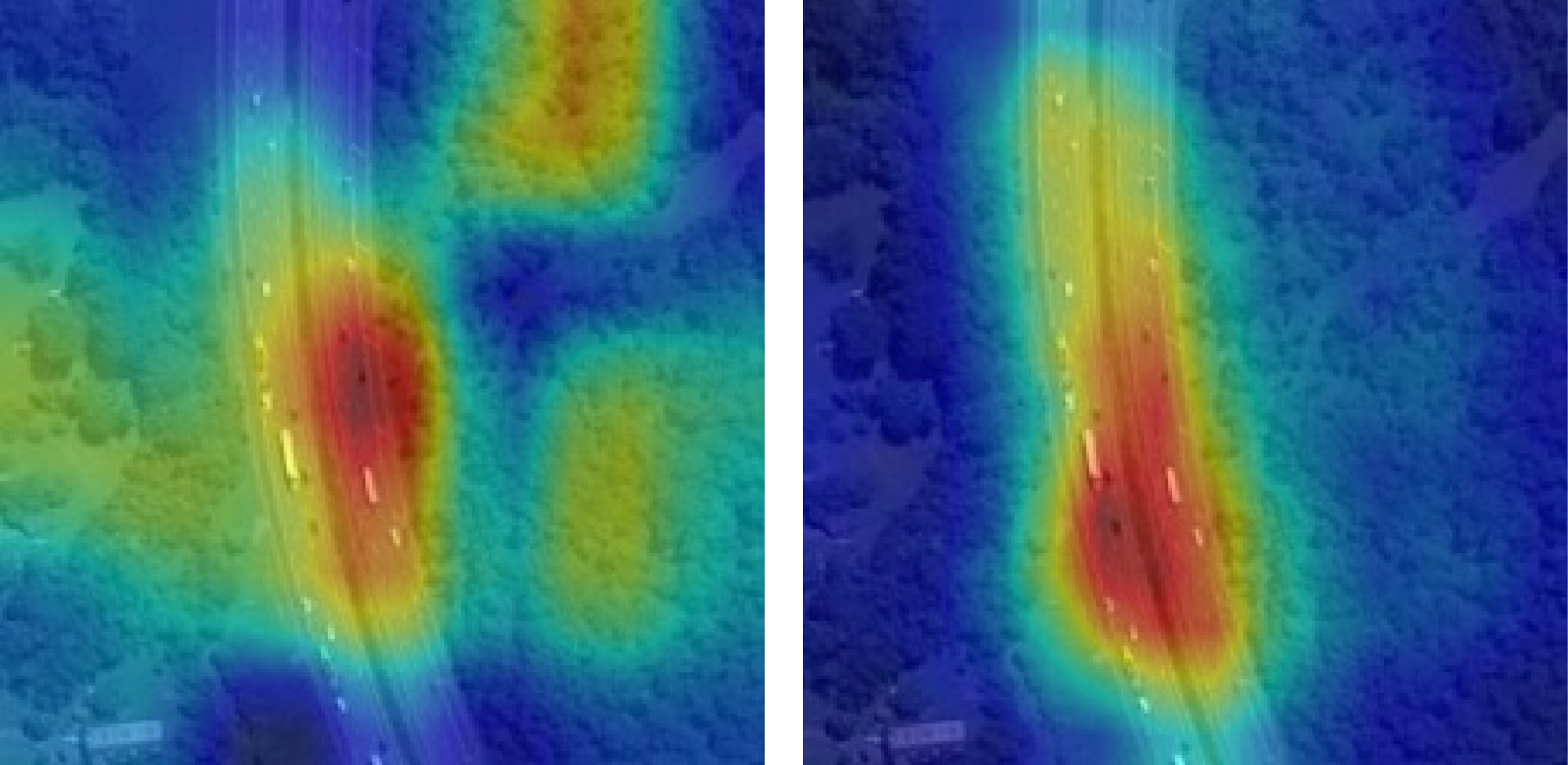}
        \caption{“Freeway” scene (NWPU-RESISC45)}
        \label{fig:cam_island}
    \end{subfigure}

    \caption{
CAM \cite{selvaraju2017grad} comparison: ResNet-50 (left) vs. Ours (right).
    }
    \label{fig:cam_comparison}
\end{figure*}

\begin{table}[t]
    \centering
    \scriptsize % 如果空间紧张可以保留，如果不紧张可以改为 \small
    \caption{Impact of architecture, fusion strategy, and classifier on aid.}
    \label{tab:ablation_comprehensive}
    \begin{tabular}{l ccc} % 去掉了 *{\linewidth} 和 fill 自动拉伸
        \toprule
        \textbf{Variant} & \textbf{P (\%)} & \textbf{R (\%)} & \textbf{F1 (\%)} \\
        \midrule
        \textbf{Full Model} & \textbf{93.76} & \textbf{93.72} & \textbf{93.71} \\
        \midrule
        \textit{Architecture Analysis} & & & \\
        w/o Global Visual Encoder    & 92.36 & 92.34 & 92.33 \\
        w/o Local Visual Encoder     & 88.19 & 88.16 & 88.13 \\
        \midrule
        \textit{Fusion Strategy} & & & \\
        Hierarchical Fusion $\rightarrow$ Concat           & 91.77 & 91.74 & 91.72 \\
        Hierarchical Fusion $\rightarrow$ Cross Attention  & 92.38 & 92.36 & 92.33 \\
        Hierarchical Fusion $\rightarrow$ AFF\cite{9423114} & 93.15 & 93.14 & 93.10 \\
        \midrule
        \textit{Classifier Analysis} & & & \\
        w/o MoE (use MLP)               & 93.30 & 93.25 & 93.28 \\
        \bottomrule
    \end{tabular}
\end{table}
\subsection{Ablation and Discussion}
The impact of the proposed design components is analyzed on the AID dataset.

\subsubsection{Impact of Parallel Heterogeneous Encoder}
Table~\ref{tab:ablation_comprehensive} validates the necessity of our heterogeneous design. 
Removing the global visual encoder leads to a 1.38\% reduction in F1 score, confirming its specific role in capturing long-range global contexts. 
More critically, eliminating the local visual encoder causes a catastrophic 5.58\% drop (plummeting to 88.13\%). 
This sharp decline supports our architectural premise: the global visual encoder serves as a lightweight enhancer rather than a standalone backbone. Due to its shallow depth (10 layers) and low computational footprint, it lacks the independent capacity to model complex patterns and inherently relies on the robust local inductive biases of the local visual encoder. Ultimately, positioning Mamba as an auxiliary branch rather than a deep standalone backbone enables a synergy that yields a more robust representation than heavy single-paradigm architectures.

\subsubsection{Component Effectiveness}
As summarized in Table~\ref{tab:ablation_comprehensive}, we conduct a comprehensive ablation study to validate the proposed fusion strategy and classifier design. For feature fusion, replacing our hierarchical fusion with simple concatenation results in a substantial 1.99\% F1 decline, while neither the computation-heavy cross-attention (92.33\%) nor the lightweight AFF~\cite{9423114} (93.10\%) matches the full model (93.71\%). This indicates our mechanism is superior in bridging the semantic gap between local and global visual features compared to generic SOTA modules. Furthermore, replacing the MoE classifier head with a standard MLP causes a discernible performance drop, confirming that dynamic routing effectively addresses the high intra-class variance in remote sensing scenes by adaptively selecting task-specific experts.

\subsubsection{Impact of MoE Expert Count}
\begin{figure}[ht!]
    \centering
    \includegraphics[width=\columnwidth]{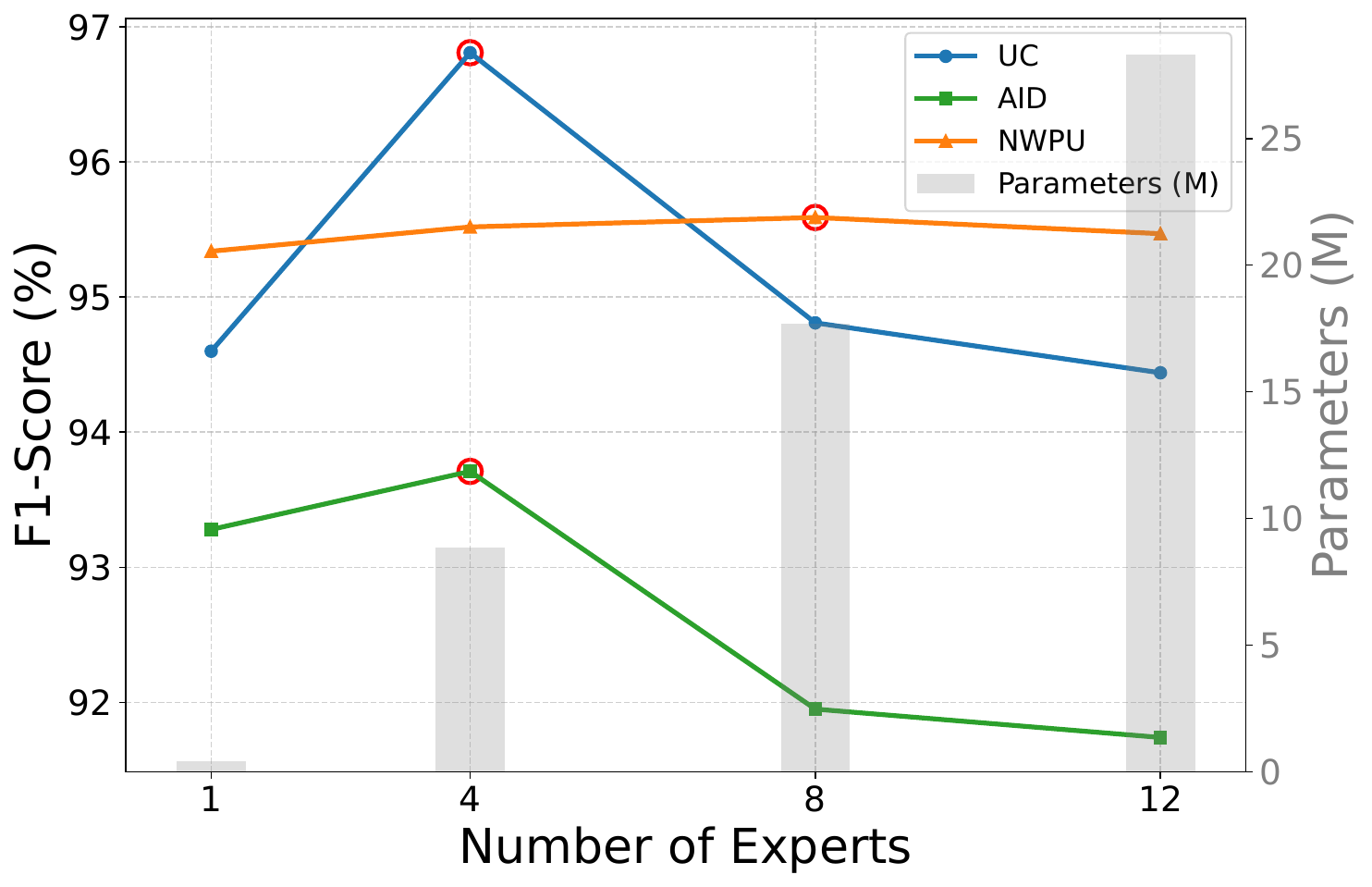}
    \caption{
Performance vs. parameters for varying experts, validating the 4-expert model.}
    \label{fig:expert_ablation}
\end{figure}

We analyze how the number of experts $N$ influences performance. Relative to the baseline ($N=1$), the 4-expert MoE achieves consistent F1 improvements by leveraging dynamic gating for adaptive feature selection. However, as depicted in \Cref{fig:expert_ablation}, scaling $N$ to 8 or 12 incurs higher computational costs (28.32M params) while performance saturates or degrades due to potential overfitting. These results confirm that $N=4$ offers the optimal trade-off, ensuring sufficient capacity for diverse scenes while maintaining parameter efficiency.

\subsection{Comparison With the State-of-the-Art}
\subsubsection{Quantitative Analysis} 
As shown in Table~\ref{tab:performance_comparison}, our model establishes new benchmarks across all datasets, surpassing the strongest competitor, CMTFNet, by 1.21\% on NWPU. Crucially, as a result of the lightweight design analyzed in the ablation study, our method achieves an optimal accuracy-efficiency trade-off (\Cref{fig:complexity_vs_accuracy}). It requires over $12\times$ fewer FLOPs (11.67G) than the 24-layer RSMamba (146.92G). These results validate that synergizing local features with a streamlined global context yields a more robust and efficient representation than heavy single-paradigm or hybrid architectures.
\subsubsection{Qualitative Analysis} 
Visualization reveals the internal mechanisms of our model. As shown in \Cref{fig:tsne_visualization}, t-SNE projections confirm that experts in the MoE module are clearly specialized for different semantic categories through structured routing. Furthermore, Grad-CAM comparisons (\Cref{fig:cam_comparison}) indicate that our model produces more focused and complete activation maps than ResNet-50. These results demonstrate that the hierarchical fusion strategy effectively concentrates attention on core semantic targets while suppressing background noise.

\section{Conclusion}
\label{sec:conclusion}
We present a parallel heterogeneous encoder that synergizes local and global visual features via a hierarchical fusion module and an adaptive MoE head. Extensive experiments across three datasets show that the proposed method achieves state-of-the-art performance with superior computational efficiency, significantly outperforming existing single-paradigm baselines. These results validate the efficacy of our hierarchical fusion strategy. Future work will extend this framework to remote sensing object detection and semantic segmentation tasks.

\bibliographystyle{IEEEbib}
\bibliography{icme2026references}

\end{document}